\documentclass[]{antgroup}
\PassOptionsToPackage{numbers, compress}{natbib}
\usepackage{antgroup}

\usepackage{graphicx}
\usepackage{tikz}
\usepackage{todonotes}
\usepackage{multirow}
\usepackage{amsmath}
\usepackage{cleveref}
\usepackage{subcaption}
\usepackage{multicol}
\usepackage{amssymb}
\usepackage{array}
\usepackage{bm}
\usepackage{enumitem}
\usepackage{algorithm}
\usepackage{algpseudocode}
\usepackage{tabularx}
\usepackage{booktabs} 
\usepackage{bbm}
\usepackage{makecell}
\usepackage{wrapfig}
\usepackage{colortbl}
\usepackage{threeparttable}
\usepackage{fontawesome5}
\usepackage[normalem]{ulem}
\usepackage{changepage}
\useunder{\uline}{\ul}{}

\usepackage{amsmath,amsfonts,bm}

\def\eqref#1{equation~\ref{#1}}

\def\1{\bm{1}}

\DeclareMathAlphabet{\mathsfit}{\encodingdefault}{\sfdefault}{m}{sl}
\SetMathAlphabet{\mathsfit}{bold}{\encodingdefault}{\sfdefault}{bx}{n}

\newcommand*\justify{%
  \fontdimen2\font=0.4em
  \fontdimen3\font=0.2em
  \fontdimen4\font=0.1em
  \fontdimen7\font=0.1em
  \hyphenchar\font=`\-
}

\renewcommand{\texttt}[1]{%
  \begingroup
  \ttfamily
  \begingroup\lccode`~=`/\lowercase{\endgroup\def~}{/\discretionary{}{}{}}%
  \begingroup\lccode`~=`[\lowercase{\endgroup\def~}{[\discretionary{}{}{}}%
  \begingroup\lccode`~=`.\lowercase{\endgroup\def~}{.\discretionary{}{}{}}%
  \catcode`/=\active\catcode`[=\active\catcode`.=\active
  \justify\scantokens{#1\noexpand}%
  \endgroup
}

\usepackage{amsmath}
\usepackage{amssymb}
\usepackage{amsfonts}                               
\usepackage{amsthm}
\usepackage[mathcal]{eucal}
\usepackage{mathrsfs}
\usepackage{bm}                                     
\usepackage{blkarray}                               
\usepackage{nicefrac}                               

\usepackage{wrapfig}
\usepackage{graphicx}                               
\usepackage{caption}
\usepackage{cleveref}

\usepackage{tikz}                                          
\usepackage{circuitikz}
\usetikzlibrary{patterns,snakes}
\usetikzlibrary{positioning,calc,fit,decorations.pathmorphing,shapes.geometric, shapes.gates.logic.US, calc}
\usetikzlibrary{arrows,arrows.meta,decorations.markings,shapes,shapes.arrows}
\usetikzlibrary{decorations,decorations.pathreplacing}
\usetikzlibrary{backgrounds}
\usepackage{filecontents}                           
\usepackage{pgfplots}
\usepackage{pgfplotstable}
\usepgfplotslibrary{groupplots}
\usepackage{scalefnt}
\pgfplotsset{compat=newest}

\usepackage{mdframed}

\usepackage{xcolor}

\definecolor{LightGray}{gray}{0.9}

\definecolor{myPink}{HTML}{EE6B98}
\definecolor{myBlue}{HTML}{6BB8FA}
\definecolor{myPurple}{HTML}{9F63F0}
\definecolor{myBlueBase}{HTML}{147BFF}

\title{LLaDA2.1: Speeding Up Text Diffusion via Token Editing}

\author{\centering
Tiwei Bie$^1$,
Maosong Cao$^1$,
Xiang Cao$^1$,
Bingsen Chen$^1$,
Fuyuan Chen$^1$,
Kun Chen$^1$,
Lun Du$^1$,\\
Daozhuo Feng$^1$,
Haibo Feng$^{1,4}$,
Mingliang Gong$^1$,
Zhuocheng Gong$^1$,
Yanmei Gu$^1$,
Jian Guan$^1$,\\
Kaiyuan Guan$^1$,
Hongliang He$^{1,3}$,
Zenan Huang$^1$,
Juyong Jiang$^1$,
Zhonghui Jiang$^1$,
Zhenzhong Lan$^{1,3,\dag}$,\\
Chengxi Li$^1$,
Jianguo Li$^{1,\dag}$,
Zehuan Li$^1$,
Huabin Liu$^1$,
Lin Liu$^1$,
Guoshan Lu$^1$,
Yuan Lu$^1$,
Yuxin Ma$^1$,\\
Xingyu Mou$^1$,
Zhenxuan Pan$^1$,
Kaida Qiu$^1$,
Yuji Ren$^1$,
Jianfeng Tan$^1$,
Yiding Tian$^1$,
Zian Wang$^1$,\\
Lanning Wei$^1$,
Tao Wu$^1$,
Yipeng Xing$^1$,
Wentao Ye$^{1,2}$,
Liangyu Zha$^1$,
Tianze Zhang$^1$,
Xiaolu Zhang$^1$,\\
Junbo Zhao$^{1,2,\dag}$,
Da Zheng$^{1,\dag}$,
Hao Zhong$^{1,2}$,
Wanli Zhong$^{1,4}$,
Jun Zhou$^1$,
Junlin Zhou$^{1}$,
Liwang Zhu$^1$,
Muzhi Zhu$^{1,2}$,
Yihong Zhuang$^1$
}

\footnotetext{Authors are listed in alphabetical order based on last name. $\dag$ indicates tech-leaders.}

\affiliation{$^1$Ant Group, $^2$Zhejiang University, $^3$Westlake University, $^4$Southern University of Science and Technology}

\begin{document}
\maketitle

\begin{abstract}
While \textbf{LLaDA2.0} showcased the scaling potential of 100B-level block-diffusion models and their inherent parallelization, the delicate equilibrium between decoding speed and generation quality has remained an elusive frontier.
Today, we unveil \textbf{LLaDA2.1}, a paradigm shift designed to transcend this trade-off. By seamlessly weaving \textbf{Token-to-Token (T2T)} editing into the conventional \textbf{Mask-to-Token (M2T)} scheme, we introduce a joint, configurable threshold-decoding scheme. This structural innovation gives rise to two distinct personas: the \textit{\textbf{Speedy Mode (S Mode)}}, which audaciously lowers the M2T threshold to bypass traditional constraints while relying on T2T to refine the output; and the \textit{\textbf{Quality Mode (Q Mode)}}, which leans into conservative thresholds to secure superior benchmark performances with manageable efficiency degrade. 
Furthering this evolution, underpinned by an expansive context window, we implement the first large-scale \textbf{Reinforcement Learning (RL)} framework specifically tailored for dLLMs, anchored by specialized techniques for stable gradient estimation. This alignment not only sharpens reasoning precision but also elevates instruction-following fidelity, bridging the chasm between diffusion dynamics and complex human intent. 
We culminate this work by releasing \textbf{LLaDA2.1-Mini (16B)} and \textbf{LLaDA2.1-Flash (100B)}. Across 33 rigorous benchmarks, LLaDA2.1 delivers strong task performance and lightning-fast decoding speed. Despite its 100B volume, on coding tasks it attains an astounding \textbf{892 TPS} on HumanEval+, \textbf{801 TPS} on BigCodeBench, and \textbf{663 TPS} on LiveCodeBench.
\begin{figure}[h]
    \centering
    \includegraphics[width=0.9\linewidth]{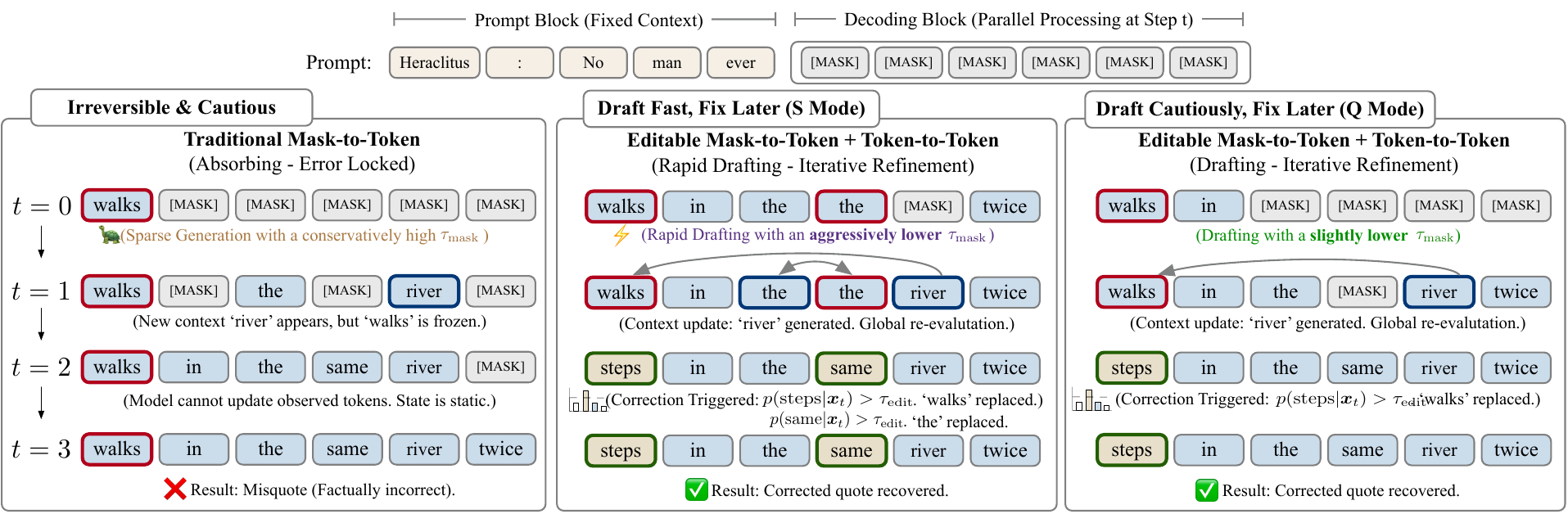}
    \caption{\textbf{Aggressive parallel drafting}, backed by retroactive correction, accelerates inference.}
    \label{fig:motivation}
\end{figure}

\end{abstract}

\newpage
\section{Introduction}

Discrete diffusion Large Language Models (dLLMs) have emerged as a compelling alternative to autoregressive generation, offering the potential for non-monotonic reasoning and parallel decoding. However, the standard absorbing-state framework—which enforces a rigid, monotonic transition from $\text{[MASK]}$ to fixed tokens—faces inherent limitations in fidelity. As highlighted by \citet{kang_parallelbench_2025}, the independent nature of parallel decoding often amplifies token-level inconsistencies. While recent studies have attempted to mitigate this via confidence-based remasking \citep{wang_remasking_2025} or by employing external guide models \citep{lee_effective_2025}. To bridge the gap between efficient parallel generation and high-fidelity reasoning, we align with the direction of generalizing discrete diffusion beyond absorbing states \citep{rutte_generalized_2025} and propose a comprehensive framework for Editable State Evolution.

Unlike prior work such as \citet{song2025seed}, we first design a novel \textbf{Error-Correcting Editable} decoding strategy, which introduces a dynamic paradigm controlled by dual probability thresholds. This paradigm encompasses two types of operations: direct decoding from mask to token, and editing from one token to another. This strategy enables the model to directly refine its own outputs during the generation process, thereby effectively addressing the local inconsistencies commonly encountered in parallel decoding. To cultivate this editing capability, our CPT and SFT phases expose the model to both masked positions and stochastic noise, incentivizing it to not only generate new content but also identify and rectify existing errors.

\textbf{Crucially, this architecture transforms the rigid trade-off between latency and fidelity into a flexible, user-configurable continuum.} By allowing the model to retroactively correct errors, we can aggressively lower the confidence threshold for the initial Mask-to-Token (M2T) phase without collapsing the generation quality. This insight gives rise to two distinct operating personas: a \textbf{\textit{Speedy Mode (S Mode)}}, which prioritizes high-throughput generation by accepting lower-confidence tokens and relying on subsequent Token-to-Token (T2T) passes for rectification; and a \textbf{\textit{Quality Mode (Q Mode)}}, which adheres to conservative thresholds to maximize reasoning rigor. This duality demonstrates that editability is not merely a mechanism for error repair, but a fundamental lever for accelerating parallel decoding.

To further elevate the model's capabilities, we integrate a \textbf{Reinforcement Learning (RL)} stage. While recent works such as SPG \citep{wang2025spg}, TraceRL \citep{wang2025revolutionizing} and ESPO \citep{ou_principled_2025} have demonstrated the potential of RL in improving dLLMs, applying policy gradients to block-autoregressive models remains challenging due to the intractability of sequence log-likelihoods. We circumvent this by adopting an ELBO-based Block-level Policy Optimization (EBPO) framework tailored for our editable setting. 

Notice that LLaDA2.1 extends its previous version (LLaDA2.0) by prioritizing decoding versatility over mere parameter scaling or benchmark peaking. By keeping the model size constant and minimal change of training data, we prove that our novel editing scheme enables lightning-fast execution with minimal overhead. This work serves as a proof-of-concept for a new dLLM paradigm that balances high-quality generation with extreme operational efficiency.

\section{Configurable Decoding Scheme}

\vspace{5pt}
\begin{mdframed}[
    linecolor=gray!40,
    linewidth=1.5pt,
    roundcorner=5pt,
    innertopmargin=5pt,
    innerbottommargin=5pt,
    innerleftmargin=10pt,
    innerrightmargin=10pt,
    backgroundcolor=gray!5
]
During LLM decoding, \textbf{Exposure Bias}—where errors compound as the model conditions on its own imperfect predictions—is inevitable. This phenomenon is particularly severe in dLLMs due to their parallel generation nature. We observe that once such decoding errors occur, dLLMs tend to become increasingly conservative in subsequent steps, significantly slowing down the generation process. In contrast, autoregressive models exhibit lower exposure bias and can self-correct through extended chain-of-thought reasoning. To address this challenge, we introduce an \textbf{editing} operation into the decoding process, enabling the model to retrospectively correct errors introduced during parallel generation, thereby achieving a much better balance between generation speed and quality.
\end{mdframed}

Specifically, we extend standard discrete diffusion to support it. Unlike conventional absorbing-state models that enforce a rigid monotonic transition from $\text{[MASK]}$ to fixed tokens, our framework introduces a dynamic ``Draft-and-Edit" paradigm controlled by dual probability thresholds.
We formalize the state evolution by defining two active update sets at timestep $t$: the \textit{Unmasking Set} $\Gamma_t$ and the \textit{Editing Set} $\Delta_t$.

We formalize the state evolution by defining two active update sets at timestep $t$: the \textit{Unmasking Set} $\Gamma_t$ and the \textit{Editing Set} $\Delta_t$.
Let $v_t^i = \arg\max_{v} p_\theta(v | \bm{x}_t)$ be the top-candidate. The update indices are identified as:
\begin{align}
    \Gamma_t &= \left\{ i \mid x_t^i = \text{[MASK]} \text{ and } p_\theta(v_t^i|\bm{x}_t) > \tau_{\text{mask}} \right\}, \\
    \Delta_t &= \left\{ i \mid x_t^i \neq v_t^i \text{ and } p_\theta(v_t^i|\bm{x}_t) > \tau_{\text{edit}} \right\},
\end{align}
with $\tau_{\text{mask}}, \tau_{\text{edit}} \in [0, 1]$ being the confidence thresholds configuring the decoding dynamics. The transition operator then applies the updates strictly on the union of these sets:
\begin{equation}
    x_{t-1}^i = 
    \begin{cases} 
        v_t^i & \text{if } i \in \Gamma_t \cup \Delta_t, \\
        x_t^i & \text{otherwise}.
    \end{cases}
    \label{eq:set_transition}
\end{equation}

\section{Training Paradigm}

\begin{figure}[t]
    \centering
    \includegraphics[width=\linewidth]{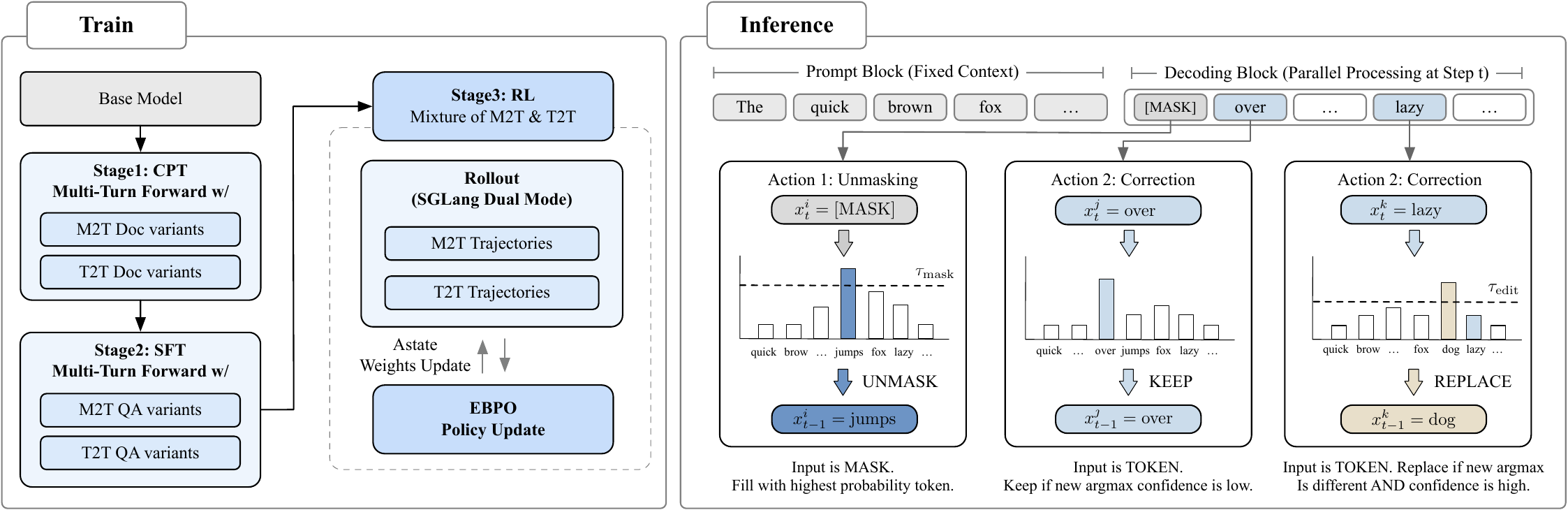}
    \caption{Overview of training \& inference framework of LLaDA2.1}
    \label{fig:framework}
\end{figure}

\subsection{Training Alignment for ``Draft-and-Edit"}
To align the model with the ``Draft-and-Edit" inference paradigm and mitigate the \emph{Exposure Bias} inherent in standard mask-based training, we employ a unified \textbf{Mixture of M2T and T2T} objective. This objective is applied throughout both the Continual Pre-Training (CPT) and Supervised Finetuning (SFT) stages.

This dual-stream training objective enables the model to develop two complementary capabilities fundamental to our framework:

\begin{itemize} 
\item \textbf{Drafting Stream (Mask-to-Token):} 
The model learns to predict the correct token at each masked position to generate initial content, establishing the foundational drafting capability. 

\item \textbf{Editing Stream (Token-to-Token):} 
The model learns to recover original tokens from random noise perturbations (rectifying errors), equipping it with the ability to identify and rewrite artifacts. 
\end{itemize}

By consistently applying this dual-stream supervision from CPT through SFT, we ensure that LLaDA2.1 is fundamentally conditioned to function as both a fast drafter and a precise editor within a single parameter space. Additionally, we employ a Multi-turn Forward (MTF) data augmentation technique, by exposing the model to a wider variety of editing scenarios, enhance the model's editing capabilities.
\subsection{Reinforcement Learning Training}

The application of policy gradient methods to diffusion models faces a fundamental hurdle: the intractability of the sequence-level log-likelihood, $\log \pi_{\theta}(\bm{x})$, which is essential for computing policy updates. While prior works have explored various approximations, they have historically struggled with high variance and prohibitive computational costs, limiting RL to small-scale experiments~\citep{wang2025revolutionizing,ou_principled_2025,wang2025spg}.
We overcome this bottleneck by synthesizing \textbf{ELBO-based Block-level Policy Optimization (EBPO)} with robust infrastructure optimizations. By utilizing the Evidence Lower Bound (ELBO) as a principled proxy for exact likelihood and implementing \textbf{Vectorized Likelihood Estimation}~\citep{Blockdiffusion2025} to parallelize bound computation, we achieve orders-of-magnitude acceleration. This integration allows us to scale dLLMs RL to unprecedented context lengths and training magnitudes, establishing a stable and efficient pipeline for post-training.

Formally, we maximize a clipped surrogate objective, where the advantage is weighted by the probability ratio $\rho$:
\begin{equation}\label{eq:ebpo_objective}
\mathcal{J}_{\text{EBPO}}(\theta) = \mathbb{E}_{\bm{x}, \bm{y} \sim \pi_{\theta_{\text{old}}}} \left[ \min \left( \rho(\bm{y}|\bm{x}) \hat{A}, \text{clip}(\rho(\bm{y}|\bm{x}), 1-\epsilon_{\text{low}}, 1+\epsilon_{\text{high}}) \hat{A} \right) \right],
\end{equation}
where $\hat{A}$ is an estimator of the advantage function at timestep $t$, quantifying the relative improvement of the chosen action over the average expectation under the current policy.
For a set of discretized timesteps $\{t_n\}_{n=1}^N$ and weights $\{w_n\}$, we construct a composite input $\bm{z}_n = \bm{y}_{t_n} \oplus \bm{y}_0$ to compute all block-conditional probabilities in parallel:
\begin{equation}
\log \rho(\bm{y}|\bm{x}) \approx \sum_{n=1}^N w_n \sum_{b=1}^B \left( \log p_\theta(\bm{y}^b \mid \bm{z}_n, \bm{x}; \mathcal{M}) - \log p_{\theta_{\text{old}}}(\bm{y}^b \mid \bm{z}_n, \bm{x}; \mathcal{M}) \right).
\end{equation}
Here, $\mathcal{M}$ denotes a Block-Causal Mask ensuring the $b$-th block attends only to valid history. By aggregating block-level contributions ($\sum_{b=1}^B$) within a single forward pass per timestep $n$, we establish a computationally tractable pipeline for scaling reinforcement learning to long-context diffusion generation.

\section{Infrastructure}
\subsection{Training Infrastructure}

\paragraph{Continued Pre-Training and Supervised Fine-Tuning}
For both continued pre-training (CPT) and supervised fine-tuning (SFT), we adopt the same training infrastructure as LLaDA2.0~\citep{bie_llada20_2025}, leveraging dFactory~\citep{dfactory}, which provides efficient training recipes specifically designed for dLLMs, except that we introduce a dedicated optimized implementation for the multi-turn forward (MTF) stage. 

\paragraph{RL Training}
To enable effective policy optimization for dLLMs, we extend the AReaL framework~\citep{fu2025areal, mei2025real} by developing specialized likelihood estimation and advantage estimation protocols that leverage diffusion sampling, explicitly supporting both T2T and M2T modes. This workflow is powered by ASystem~\citep{lingteam2025stepevolvesscalingreinforcement} for distributed orchestration and utilizes a customized version of SGLang~\citep{antgroupdeepxputeamPowerDiffusionLLMs} as the dedicated rollout engine.

\subsection{Inference Infrastructure}
We use a customized version of SGLang~\citep{antgroupdeepxputeamPowerDiffusionLLMs} for inference. To further accelerate the inference speed, we integrate Alpha-MoE~\citep{alphamoe}, a MoE megakernel that combines the two FusedMoE computations into one kernel, and adopt per-block FP8 quantization to balance the inference speed and model accuracy. To accelerate inference on long-context sequences, we adopt block-wise causal masked attention, allowing the KV cache for the entire long context to be computed in a single forward pass. We further enable radix caching and batching support for block diffusion LLMs in SGLang.

\subsection{Decoding Algorithm at Inference}
In the inference stage, we adopt a decoding algorithm that combines Threshold Decoding~\citep{ma2025dinfer} with an explicit editing mechanism. In the basic setting, decoding and editing are performed within a single block: tokens are generated under a threshold-based constraint, and local edits are applied to revise intermediate outputs before the block is finalized.

Beyond single-block editing, we further introduce a \textbf{Multiple Block Editing (MBE)} mechanism. MBE allows the model to revisit and revise previously generated blocks based on the content of newly decoded blocks.
\section{Evaluation}

To comprehensively evaluate the quality of instruction-tuned models, we employ a diverse suite of benchmarks categorized into five dimensions:

\begin{itemize}
    \item \textbf{Knowledge}: MMLU-Pro \citep{wang2024mmlupro}, GPQA-Diamond \citep{rein2024gpqa}, C-Eval \citep{huang2023ceval}, PHYBench \citep{qiu2025phybench}, TriviaQA \citep{joshi2017triviaqa}
    \item \textbf{Reasoning}: SQuAD 2.0 \citep{rajpurkar2018know}, DROP \citep{dua2019drop}, KOR-Bench \citep{ma2024kor}, HellaSwag \citep{zellers2019hellaswag}, BIG-Bench Hard \citep{suzgun2023challenging}, BIG-Bench Extra Hard \citep{kazemi2025big}, MuSR \citep{sprague2023musr}, ZebraLogic \citep{lin2025zebralogic}, PrOntoQA \citep{saparov2022language}, PIQA \citep{bisk2020piqa}, OCNLI \citep{hu2020ocnli}, BIG-Bench Hard-CN \citep{opencompass}
    \item \textbf{Coding}: CRUXEval \citep{gu2024cruxeval}, MultiPL-E \citep{cassano2023multiple}, BigCodeBench \citep{zhuo2024bigcodebench}, LiveCodeBench \citep{jain2024livecodebench}, Spider \citep{yu2018spider}, BIRD \citep{li2023can}, HumanEval+ \citep{liu2023your}, MBPP+ \citep{liu2023your}
    \item \textbf{Math}: OlympiadBench \citep{he2024olympiadbench}, AIME 2025 \citep{aime2025aime}, Omni-MATH \citep{gao2024omni}, GSM-Plus \citep{li2024gsm}, CMATH \citep{wei2023cmath}
    \item \textbf{Agent \& Alignment}: BFCL \citep{patil2025bfcl}, IFEval \citep{zhou2023ifeval}, Nexus Function Calling Benchmark \citep{nexusraven}
\end{itemize}

We report the comparative scores and TPF (tokens per forward) of LLaDA2.1-flash and LLaDA2.1-mini against other models in Tables ~\ref{tab:flash_benchmark} and ~\ref{tab:mini_benchmark}, respectively. From the results, we observe that LLaDA2.1’s scores under \textit{S Mode} decrease compared to LLaDA2.0, but a substantial improvement in TPF is achieved.  While under \textit{Q Mode}, LLaDA2.1 surpasses the results of LLaDA2.0 on both mini and flash model.

In Table~\ref{tab:flash_mini_quant}, we focus on showcasing the speed performance of LLaDA2.1 in \textit{S Mode}. It can be observed that LLaDA2.1 exhibits significant speed variations across different domains, being highest in the code domain and lowest in instruction following. Specifically, after quantization, LLaDA2.1-flash achieves a peak TPS of 891.74 on HumanEval+, while LLaDA2.1-mini reaches 1586.93 in peak TPS, demonstrating significant speed advantages.

\begin{figure}[h]
    \centering
    \includegraphics[width=0.9\linewidth]{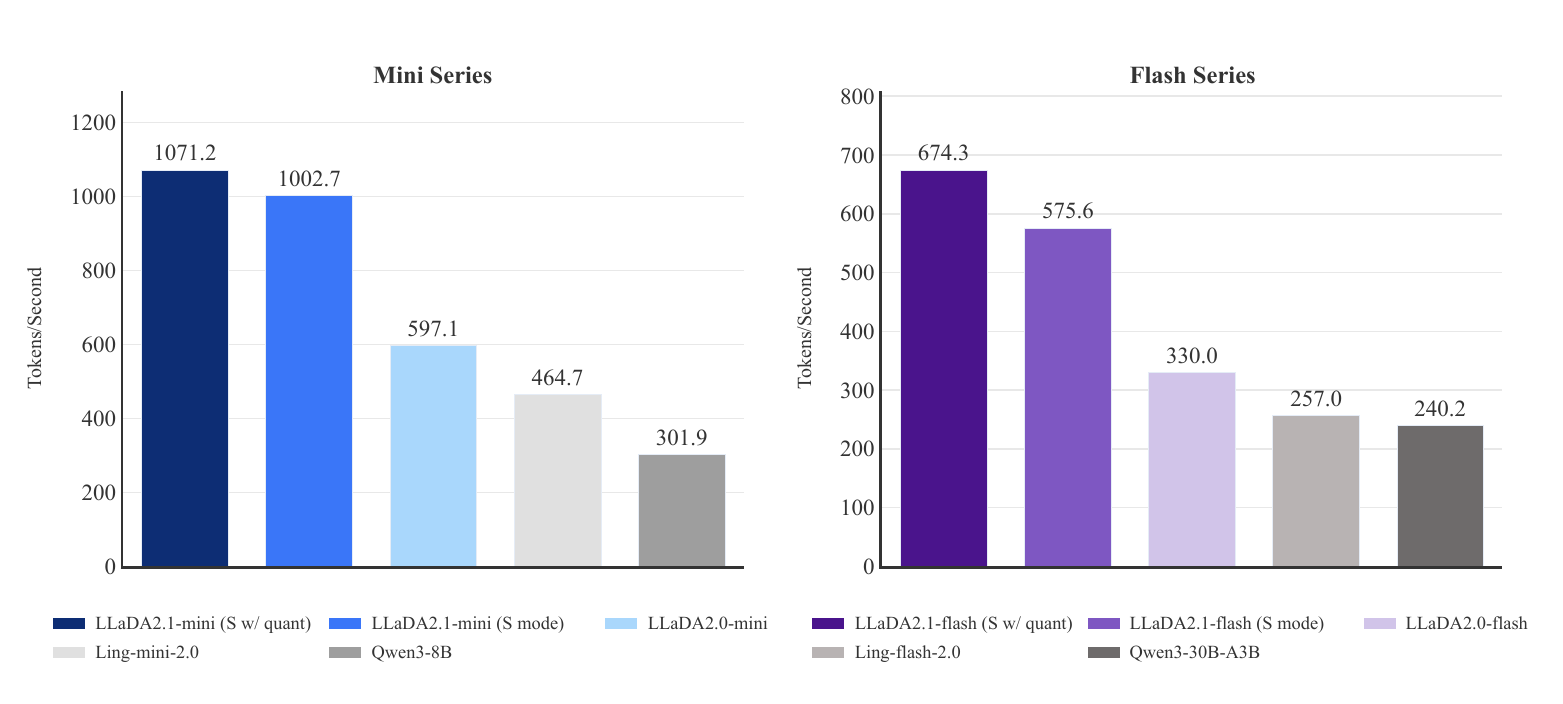}
    \caption{Throughput (TPS) comparison on nine benchmarks, consistent with the evaluation settings in Table \ref{tab:flash_mini_quant}, for LLaDA2.1 variants against LLaDA2.0, Ling, and Qwen3 across the mini (left) and flash (right) series.}
    \label{fig:tps}
\end{figure}

As shown in Table~\ref{tab:flash_mini_mbe}, under the same \textit{S Mode} setting, Multi-Block Editing (MBE) yields consistent performance improvements across benchmarks for both Flash and Mini variants, at the cost of a modest reduction in throughput. The gains are particularly evident on reasoning and coding tasks, indicating that iterative cross-block refinement effectively corrects local errors and improves global consistency without substantially compromising decoding efficiency.

Figure~\ref{fig:tps} further illustrates the throughput (in terms of token per sec) comparison  of LLaDA 2.1 variants against LLaDA 2.0, Ling, and Qwen-3 across 5 different benchmark domains as shown in Table~\ref{tab:flash_mini_quant}. 
This comparison spotlights LLaDA-2.1 (S Mode)’s striking speed advantage: it achieves dramatically faster inference while sacrificing only a negligible sliver of output quality.

\begin{table}[htbp]
\centering
\caption{Benchmark Performance of LLaDA2.1-flash, comparing with several baseline models. For diffusion language model, we report its scores across each benchmark along with its TPF (tokens per forward); for AR model, we report its scores only, as its TPF is inherently equal to 1.}
\label{tab:flash_benchmark}
\resizebox{\textwidth}{!}{%
\begin{tabular}{lcc|ccc}
\toprule
\textbf{Benchmark} 
& \makecell[b]{\textbf{Qwen3-30B-} \\ \textbf{A3B-Inst-2507} \\ \scriptsize (Score)}
& \makecell[b]{\textbf{Ling-flash-2.0} \\ \textbf{}\\ \scriptsize (Score)}
& \makecell[b]{\textbf{LLaDA2.0-flash} \\ \textbf{}\textbf{} \\ \scriptsize (Score $\mid$ TPF)}
& \makecell[b]{\textbf{LLaDA2.1-flash} \\ \textbf{(S Mode)}\\ \scriptsize (Score $\mid$ TPF)}
& \makecell[b]{\textbf{LLaDA2.1-flash} \\ \textbf{(Q Mode)}\\ \scriptsize (Score $\mid$ TPF)} 
\\

\midrule
\textbf{Average}& 73.09 & 71.52 & 72.43 $\mid$ 3.08 & 72.34 $\mid$ 5.93  & 73.54 $\mid$ 3.64 \\
\midrule
\multicolumn{6}{c}{\textbf{Knowledge}} \\
\midrule
GPQA & 54.14 & 69.16 & 62.31 $\mid$ 3.29 & 66.67 $\mid$ 3.95  & 67.30 $\mid$ 2.37 \\
MMLU-Pro & 74.21 & 77.55 & 74.79 $\mid$ 2.36 & 75.31 $\mid$ 4.43  & 76.59 $\mid$ 2.62 \\
C-EVAL & 88.12 & 87.54 & 85.21 $\mid$ 1.90 & 86.93 $\mid$ 2.71  & 86.71 $\mid$ 1.75 \\
PHYBench & 29.84 & 27.67 & 30.06 $\mid$ 2.70 & 26.04 $\mid$ 4.10  & 28.23 $\mid$ 2.66 \\
TriviaQA & 65.61 & 69.76 & 66.88 $\mid$ 1.94 & 72.55 $\mid$ 4.30  & 72.93 $\mid$ 2.92 \\
\midrule
\multicolumn{6}{c}{\textbf{Reasoning}} \\
\midrule
BIG-Bench Hard & 85.54 & 89.36 & 86.75 $\mid$ 2.66 & 87.82 $\mid$ 5.61  & 88.69 $\mid$ 3.28 \\
BIG-Bench Extra Hard & 37.80 & 23.24 & 27.86 $\mid$ 4.60 & 33.51 $\mid$ 5.04 &  35.77 $\mid$ 3.17 \\
bbh-zh & 86.18 & 75.09 & 87.52 $\mid$ 3.21 & 82.55 $\mid$ 5.78  & 86.23 $\mid$ 3.77 \\
MuSR & 79.15 & 82.72 & 80.48 $\mid$ 1.70 & 80.10 $\mid$ 2.90 & 79.84 $\mid$ 1.85 \\
ZebraLogic & 90.97 & 87.60 & 82.30 $\mid$ 2.74 & 84.20 $\mid$ 5.80  & 88.90 $\mid$ 3.26 \\
PrOntoQA & 97.12 & 97.88 & 96.50 $\mid$ 2.64 & 95.00 $\mid$ 9.23  & 97.00 $\mid$ 5.73 \\
PIQA & 91.57 & 91.95 & 92.76 $\mid$ 1.43 & 92.44 $\mid$ 2.38  & 92.17 $\mid$ 1.44 \\
OCNLI & 71.59 & 65.36 & 71.63 $\mid$ 1.09 & 72.17 $\mid$ 1.83  & 72.75 $\mid$ 1.32 \\
HellaSwag & 86.31 & 81.59 & 84.97 $\mid$ 1.26 & 85.60 $\mid$ 2.31  & 85.31 $\mid$ 1.51 \\
KOR-Bench & 69.20 & 69.44 & 63.04 $\mid$ 3.44 & 62.80 $\mid$ 4.97  & 65.12 $\mid$ 2.77 \\
DROP & 87.57 & 88.32 & 87.90 $\mid$ 2.26 & 87.55 $\mid$ 5.40  & 87.86 $\mid$ 2.53 \\
SQuAD 2.0 & 89.51 & 81.32 & 90.00 $\mid$ 3.10 & 90.65 $\mid$ 5.01  & 90.80 $\mid$ 3.90 \\
\midrule
\multicolumn{6}{c}{\textbf{Coding}} \\
\midrule
LiveCodeBench & 46.42 & 52.48 & 42.51 $\mid$ 4.23 & 44.05 $\mid$ 6.48  & 45.37 $\mid$ 3.80 \\
CRUXEval-O & 86.75 & 82.75 & 85.12 $\mid$ 3.21 & 85.25 $\mid$ 6.54  & 87.50 $\mid$ 3.80 \\
MBPP+ & 78.21 & 80.89 & 79.37 $\mid$ 4.02 & 76.72 $\mid$ 10.43  & 77.25 $\mid$ 5.96 \\
HumanEval+ & 87.88 & 87.58 & 88.41 $\mid$ 6.45 & 89.63 $\mid$ 13.81  & 89.63 $\mid$ 9.18 \\
MultiPL-E & 70.67 & 65.76 & 74.87 $\mid$ 3.14 & 70.89 $\mid$ 7.77  & 73.34 $\mid$ 4.33 \\
BigCodeBench-Full & 41.49 & 40.70 & 41.58 $\mid$ 3.33 & 37.11 $\mid$ 8.51 & 39.21 $\mid$ 4.70 \\
BIRD-SQL & 47.75 & 47.49 & 45.76 $\mid$ 2.16 & 42.18 $\mid$ 5.09 &  44.04 $\mid$ 2.95 \\
Spider & 81.79 & 80.58 & 82.49 $\mid$ 4.42 & 79.18 $\mid$ 8.74 &  81.04 $\mid$ 5.70 \\
\midrule
\multicolumn{6}{c}{\textbf{Math}} \\
\midrule
AIME 2025 & 61.88 & 55.89 & 60.00 $\mid$ 4.57 & 63.33 $\mid$ 5.36 &  63.33 $\mid$ 3.46 \\
OlympiadBench & 77.59 & 76.19 & 74.07 $\mid$ 3.70 & 75.85 $\mid$ 6.46 &  76.59 $\mid$ 3.81 \\
GSM-Plus & 89.41 & 89.71 & 89.74 $\mid$ 2.68 & 89.23 $\mid$ 7.14 &  89.69 $\mid$ 3.83 \\
CMATH & 96.58 & 96.52 & 96.90 $\mid$ 2.17 & 96.54 $\mid$ 4.84 &  96.63 $\mid$ 2.65 \\
Omni-MATH & 54.00 & 53.00 & 50.30 $\mid$ 3.39 & 52.30 $\mid$ 6.01 &  54.10 $\mid$ 3.50 \\
\midrule
\multicolumn{6}{c}{\textbf{Agent \& Alignment}} \\
\midrule
IFEval-strict-prompt & 83.73 & 81.15 & 82.62 $\mid$ 1.47 & 83.36 $\mid$ 2.24 & 83.55 $\mid$ 1.41 \\
BFCL v3 & 73.41 & 67.69 & 74.94 $\mid$ 4.87 & 74.86 $\mid$ 9.24 &  75.61 $\mid$ 6.76 \\
Nexus FC & 49.93 & 36.25 & 50.45 $\mid$ 5.53 & 44.83 $\mid$ 11.29 &  47.65 $\mid$ 7.38 \\
\bottomrule
\end{tabular}

}
\end{table}

\begin{table}[htbp]
\centering
\caption{Benchmark Performance of LLaDA2.0-mini, comparing with several baseline models. For diffusion language model, we report its scores across each benchmark along with its TPF (tokens per forward); for AR model, we report its scores only, as its TPF is inherently equal to 1.}
\label{tab:mini_benchmark}
\footnotesize
\resizebox{\textwidth}{!}{%
\begin{tabular}{lcc|ccc}
\toprule
\textbf{Benchmark} 
& \makecell[b]{\textbf{Qwen3-8B} \\ \textbf{(no\_think)} \\ \scriptsize (Score)} 
& \makecell[b]{\textbf{Ling-mini-2.0} \\ \textbf{} \\ \scriptsize (Score)}
& \makecell[b]{\textbf{LLaDA2.0-mini} \\ \textbf{} \\ \scriptsize (Score $\mid$ TPF)}
& \makecell[b]{\textbf{LLaDA2.1-mini} \\ \textbf{(S Mode)} \\ \scriptsize (Score $\mid$ TPF)}
& \makecell[b]{\textbf{LLaDA2.1-mini} \\ \textbf{(Q Mode)} \\ \scriptsize (Score $\mid$ TPF)} 
\\
\midrule							
\textbf{Average} & 61.59 & 64.72 & 63.39 $\mid$ 2.60 & 62.07 $\mid$ 5.34 & 63.90 $\mid$ 3.12 \\
\midrule
\multicolumn{6}{c}{\textbf{Knowledge}} \\
\midrule
GPQA & 48.01 & 59.41 & 47.76 $\mid$ 2.73 & 48.36 $\mid$ 3.62 &  53.28 $\mid$ 2.12 \\
MMLU-Pro & 65.83 & 67.18 & 64.27 $\mid$ 2.15 & 63.42 $\mid$ 4.22 &  64.84 $\mid$ 2.41 \\
C-EVAL & 80.60 & 82.17 & 81.80 $\mid$ 1.78 & 78.40 $\mid$ 3.39 &  78.59 $\mid$ 1.91 \\
PHYBench & 9.76 & 14.59 & 11.70 $\mid$ 2.48 & 12.75 $\mid$ 4.41 &  13.05 $\mid$ 2.52 \\
TriviaQA & 52.51 & 55.63 & 51.33 $\mid$ 1.54 & 53.33 $\mid$ 3.21 &  54.24 $\mid$ 2.02 \\
\midrule
\multicolumn{6}{c}{\textbf{Reasoning}} \\
\midrule
BIG-Bench Hard & 79.48 & 83.70 & 78.21 $\mid$ 2.36 & 78.42 $\mid$ 5.02 &  80.58 $\mid$ 2.86 \\
BIG-Bench Extra Hard & 18.27 & 14.81 & 16.47 $\mid$ 2.03 & 15.30 $\mid$ 3.19 &  15.78 $\mid$ 1.66 \\
bbh-zh & 80.09 & 66.11 & 75.75 $\mid$ 2.77 & 67.65 $\mid$ 3.89 &  70.40 $\mid$ 2.35 \\
MuSR & 70.02 & 71.36 & 71.48 $\mid$ 1.45 & 70.43 $\mid$ 2.48 &  71.89 $\mid$ 1.56 \\
ZebraLogic & 37.48 & 79.85 & 64.20 $\mid$ 2.30 & 68.50 $\mid$ 5.38 &  77.10 $\mid$ 2.93 \\
PrOntoQA & 93.12 & 96.06 & 86.00 $\mid$ 2.36 & 87.50 $\mid$ 4.86 &  84.50 $\mid$ 2.73 \\
PIQA & 88.30 & 87.54 & 86.51 $\mid$ 1.45 & 84.87 $\mid$ 2.59 &  86.89 $\mid$ 1.45 \\
OCNLI & 61.49 & 60.17 & 64.51 $\mid$ 4.06 & 61.02 $\mid$ 1.78 &  61.59 $\mid$ 1.23 \\
HellaSwag & 79.56 & 69.02 & 79.01 $\mid$ 1.50 & 75.71 $\mid$ 2.39 &  76.19 $\mid$ 1.49 \\
KOR-Bench & 54.96 & 63.20 & 49.92 $\mid$ 2.45 & 46.64 $\mid$ 4.28 &  48.00 $\mid$ 2.35 \\
DROP & 84.56 & 78.80 & 81.91 $\mid$ 2.02 & 81.55 $\mid$ 5.84 &  82.37 $\mid$ 2.87 \\
SQuAD 2.0 & 85.21 & 75.56 & 86.50 $\mid$ 2.47 & 84.51 $\mid$ 4.33 &  85.13 $\mid$ 3.09 \\
\midrule
\multicolumn{6}{c}{\textbf{Coding}} \\
\midrule
LiveCodeBench & 26.76 & 42.29 & 31.83 $\mid$ 3.34 & 28.85 $\mid$ 6.42 &  30.40 $\mid$ 3.63 \\
CRUXEval-O & 74.06 & 76.12 & 71.62 $\mid$ 2.78 & 70.62 $\mid$ 5.85 &  73.75 $\mid$ 3.35 \\
MBPP+ & 72.69 & 77.25 & 78.24 $\mid$ 3.43 & 73.28 $\mid$ 10.59 &  74.07 $\mid$ 6.30 \\
HumanEval+ & 79.50 & 80.03 & 81.40 $\mid$ 5.16 & 80.49 $\mid$ 12.32 &  82.93 $\mid$ 7.77 \\
MultiPL-E & 61.70 & 67.09 & 67.46 $\mid$ 2.78 & 64.16 $\mid$ 7.23 &  67.17 $\mid$ 4.01 \\
BigCodeBench-Full & 36.05 & 35.00 & 32.89 $\mid$ 2.87 & 30.18 $\mid$ 7.33 &  34.39 $\mid$ 4.09 \\
BIRD-SQL & 36.11 & 39.67 & 39.34 $\mid$ 1.96 & 37.32 $\mid$ 4.48 &  38.40 $\mid$ 2.42 \\
Spider & 72.80 & 76.43 & 76.76 $\mid$ 3.93 & 75.78 $\mid$ 7.98 &  77.55 $\mid$ 5.48 \\
\midrule
\multicolumn{6}{c}{\textbf{Math}} \\
\midrule
AIME 2025 & 22.08 & 47.66 & 36.67 $\mid$ 2.41 & 36.67 $\mid$ 6.34 &  43.33 $\mid$ 3.29 \\
OlympiadBench & 55.33 & 72.30 & 67.70 $\mid$ 2.63 & 64.30 $\mid$ 7.08 &  66.67 $\mid$ 3.99 \\
GSM-Plus & 85.56 & 87.18 & 86.50 $\mid$ 2.41 & 85.88 $\mid$ 6.82 &  86.55 $\mid$ 3.69 \\
CMATH & 95.42 & 96.40 & 95.72 $\mid$ 1.98 & 95.63 $\mid$ 4.94 &  94.99 $\mid$ 2.56 \\
Omni-MATH & 33.20 & 48.80 & 41.70 $\mid$ 2.57 & 41.70 $\mid$ 6.41 &  43.60 $\mid$ 3.56 \\
\midrule
\multicolumn{6}{c}{\textbf{Agent \& Alignment}} \\
\midrule
IFEval-strict-prompt & 84.29 & 76.16 & 80.78 $\mid$ 1.24 & 81.33 $\mid$ 1.83 &  83.18 $\mid$ 1.25 \\
BFCL v3 & 70.12 & 53.75 & 70.72 $\mid$ 4.26 & 72.06 $\mid$ 7.39 &  73.61 $\mid$ 5.14 \\
Nexus FC & 37.71 & 34.38 & 35.18 $\mid$ 4.06 & 31.59 $\mid$ 8.27 &  33.69 $\mid$ 4.91 \\
\bottomrule
\end{tabular}%
}
\end{table}

\newcommand{\tpsdelta}[2]{%
  \makecell{%
    \begin{tabular}{@{}p{3em}|p{3em}@{}}
    #1 & #2
    \end{tabular}}}

\begin{table}[t]
\caption{Throughput (TPS) and relative score changes of Flash and Mini variants across benchmarks.
For each model family, the w/o Quant setting serves as the baseline.
Cells under w/ Quant are vertically split into \textit{TPS $\mid$ $\Delta$Score}.}
\centering
\footnotesize
\setlength{\tabcolsep}{4pt}
\begin{tabular}{l l cc cc}
\toprule
\textbf{Category}
& \textbf{Benchmark}
& \multicolumn{2}{c}{\textbf{LLaDA2.1-flash}}
& \multicolumn{2}{c}{\textbf{LLaDA2.1-mini}} \\
\cmidrule(lr){3-4}
\cmidrule(lr){5-6}
& & \makecell{\textbf{w/o Quant}\\TPS} & \makecell{\textbf{w/ Quant}\\TPS $\mid$ $\Delta$Score}
& \makecell{\textbf{w/o Quant}\\TPS} & \makecell{\textbf{w/ Quant}\\TPS $\mid$ $\Delta$Score}\\
\midrule

\multirow{5}{*}{Coding}
& HumanEval+ &
746.66 &
\tpsdelta{891.74}{-3.04} &
1496.67 &
\tpsdelta{1586.93}{-0.61} \\

& MBPP+ &
639.47 & \tpsdelta{761.38}{-1.85} &
1286.96 &
\tpsdelta{1303.96}{+1.85} \\

& CRUXEval-O &
550.09 &\tpsdelta{645.72}{-0.24} &
980.82 & \tpsdelta{1063.94}{-1.00} \\

& BigCodeBench-Full &
691.14 &
\tpsdelta{801.48}{+1.06} &
1220.40 &
\tpsdelta{1307.45}{-0.09} \\

& LiveCodeBench &
571.60 &
\tpsdelta{663.39}{-1.76} &
1015.82 &
\tpsdelta{1102.92}{+1.98} \\
\midrule

Math
& GSM-Plus &
574.65 &
\tpsdelta{667.07}{-0.03} &
1080.51 &
\tpsdelta{1186.18}{-0.30} \\
\midrule

Knowledge
& GPQA-Diamond &
416.92 &
\tpsdelta{477.79}{-0.64} &
724.30 &
\tpsdelta{784.62}{-1.64} \\
\midrule

\makecell[l]{Instruction\\Following}
& IFEval &
219.37 &
\tpsdelta{248.25}{+1.48} &
338.58 &
\tpsdelta{365.52}{-1.29} \\
\midrule

Reasoning
& PrOntoQA &
770.88 &
\tpsdelta{912.16}{-1.00} &
880.19 &
\tpsdelta{938.93}{-1.50} \\
\bottomrule
\end{tabular}
\label{tab:flash_mini_quant}
\end{table}

\newcommand{\tpfdelta}[2]{%
  \makecell{%
    \begin{tabular}{@{}p{2.5em}|p{3em}@{}}
    #1 & #2
    \end{tabular}}}
\begin{table}[t]
\caption{
Performance comparison of LLaDA2.1-flash and Mini variants with and without Multi-Block Editing (MBE) across benchmarks.
Each cell reports \textit{Score $\mid$ TPF}.
}
\centering
\footnotesize
\setlength{\tabcolsep}{4pt}
\begin{tabular}{l l cc cc}
\toprule
\textbf{Category} & \textbf{Benchmark}
& \multicolumn{2}{c}{\textbf{LLaDA2.1-flash}}
& \multicolumn{2}{c}{\textbf{LLaDA2.1-mini}} \\
\cmidrule(lr){3-4} \cmidrule(lr){5-6}
& 
& \makecell{\textbf{w/o MBE}\\\tpfdelta{Score}{TPF}}
& \makecell{\textbf{w/ MBE}\\\tpfdelta{Score}{TPF}}
& \makecell{\textbf{w/o MBE}\\\tpfdelta{Score}{TPF}}
& \makecell{\textbf{w/ MBE}\\\tpfdelta{Score}{TPF}} \\
\midrule

\multirow{2}{*}{Knowledge}
& MMLU-Pro
& \tpfdelta{75.31}{4.43}
& \tpfdelta{75.90}{3.88}
& \tpfdelta{63.42}{4.22}
& \tpfdelta{63.10}{3.66} \\
& TriviaQA
& \tpfdelta{72.55}{4.30}
& \tpfdelta{72.45}{4.28}
& \tpfdelta{53.33}{3.21}
& \tpfdelta{53.41}{3.14} \\

\midrule
\multirow{2}{*}{Reasoning}
& bbh-zh
& \tpfdelta{82.55}{5.78}
& \tpfdelta{83.21}{4.85}
& \tpfdelta{67.65}{3.89}
& \tpfdelta{67.94}{3.41} \\
& ZebraLogic
& \tpfdelta{84.20}{5.80}
& \tpfdelta{88.20}{5.03}
& \tpfdelta{68.50}{5.38}
& \tpfdelta{70.00}{4.62} \\

\midrule
\multirow{4}{*}{Coding}
& LiveCodeBench
& \tpfdelta{44.05}{6.48}
& \tpfdelta{46.48}{5.62}
& \tpfdelta{28.85}{6.42}
& \tpfdelta{29.74}{5.44} \\
& CRUXEval-O
& \tpfdelta{85.25}{6.54}
& \tpfdelta{87.00}{5.62}
& \tpfdelta{70.62}{5.85}
& \tpfdelta{70.62}{5.02} \\
& BigCodeBench-Full
& \tpfdelta{37.11}{8.51}
& \tpfdelta{39.30}{7.00}
& \tpfdelta{30.18}{7.33}
& \tpfdelta{30.70}{6.05} \\
& Spider
& \tpfdelta{79.18}{8.74}
& \tpfdelta{80.58}{8.33}
& \tpfdelta{75.78}{7.98}
& \tpfdelta{76.67}{7.59} \\

\midrule
Math
& AIME 2025
& \tpfdelta{63.33}{5.36}
& \tpfdelta{70.00}{4.71}
& \tpfdelta{36.67}{6.34}
& \tpfdelta{36.67}{5.25} \\

\midrule
Agent \& Alignment
& IFEval-strict-prompt
& \tpfdelta{83.36}{2.24}
& \tpfdelta{83.55}{2.11}
& \tpfdelta{81.33}{1.83}
& \tpfdelta{83.55}{1.70} \\

\midrule
\textbf{Average} & --
& \tpfdelta{70.69}{5.82}
& \tpfdelta{72.67}{5.14}
& \tpfdelta{57.63}{5.25}
& \tpfdelta{58.24}{4.59} \\

\bottomrule
\end{tabular}
\label{tab:flash_mini_mbe}
\end{table}

\section{Outlook and Limitation}

\paragraph{Tradeoff Between Inference Speed and Accuracy}  
While LLaDA2.1 significantly improves inference speed, a clear speed-accuracy tradeoff persists, particularly with noticeable performance differences across various domains. It is necessary to adjust threshold parameters for different domains to balance speed and accuracy. In structured-data fields such as code and math, setting \textit{S Mode} achieves high speed with little accuracy loss. However, in some general chat cases, these settings can cause undesirable output. In such cases, we recommend adjusting the parameters to \textit{Q Mode}. Our conjecture is that this pattern may be related to the model's inherent preference for structured data or the distributional characteristics of training dataset. Further validation will be conducted in our future research.

\paragraph{Editable Enhanced dLLM}
Although dLLMs inherently support high parallelism, theoretically offering speed advantages over AR models, our experimental observations show that this high parallelism also introduces a higher error rate compared to AR models. These hidden errors can reduce the model’s confidence in subsequent reasoning, ultimately slowing down the overall process. Therefore, timely editing to correct errors is essential. In our case analysis of LLaDA2.1, we observed that prompt editing corrected decoding errors, helping to maintain higher inference speeds. However, research on the editing capabilities of dLLMs is still in its early stages. We anticipate that future work, such as integrating editing into reinforcement learning, will further enhance the performance of editable dLLMs.

\paragraph{LLaDA2.1 remains in an experimental phase. Although rare, certain edge cases may occur.} Empirical observations show that aggressively lowering the masking threshold $\tau_{\text{mask}}$ can quickly generate ``rough drafts". Although the model's self-correction can partially alleviate the ``stuttering" artifacts (such as n-gram repetitions) caused by independent parallel sampling, balancing drafting speed with the quality of the initial structure remains a key operational frontier. Overall, by unifying dynamic inference, hybrid training, and principled reinforcement learning, our work establishes a solid foundation for self-correcting discrete diffusion language models.

\paragraph{Conclusion} Overall, LLaDA2.1 introduces an editing feature, which, through cumulative error correction, significantly lowered the decoding threshold of the dLLM and yielded considerable inference speed benefits. However, this model still faces many unresolved issues, and we anticipate that more powerful editable dLLMs will deliver even more unexpected and impressive results.

\bibliographystyle{antgroup}

\bibliography{ref/reference}

@string{aaai     = "Proc.~AAAI"}

@string{acl      = "Proc.~ACL"}

@string{mlsys    = "Proc.~MLSys"}

@string{test     = "Test"}

@string{arxiv    = "arXiv preprint"}

@misc{bie_llada20_2025,
    title = {{LLaDA2}.0: {Scaling} {Up} {Diffusion} {Language} {Models} to {100B}},
    shorttitle = {{LLaDA2}.0},
    url = {http://arxiv.org/abs/2512.15745},
    doi = {10.48550/arXiv.2512.15745},
    urldate = {2025-12-19},
    publisher = {arXiv},
    author = {Bie, Tiwei and Cao, Maosong and Chen, Kun and Du, Lun and Gong, Mingliang and Gong, Zhuochen and Gu, Yanmei and Hu, Jiaqi and Huang, Zenan and Lan, Zhenzhong and Li, Chengxi and Li, Chongxuan and Li, Jianguo and Li, Zehuan and Liu, Huabin and Liu, Ling and Lu, Guoshan and Lu, Xiaocheng and Ma, Yuxin and Tan, Jianfeng and Wei, Lanning and Wen, Ji-Rong and Xing, Yipeng and Zhang, Xiaolu and Zhao, Junbo and Zheng, Da and Zhou, Jun and Zhou, Junlin and Zhou, Zhanchao and Zhu, Liwang and Zhuang, Yihong},
    month = dec,
    year = {2025},
    note = {arXiv:2512.15745 [cs]},
}

@string{aaai     = "AAAI Conference on Artificial Intelligence (AAAI)"}

@string{acl      = "Annual Meeting of the Association for Computational Linguistics (ACL)"}

@string{mlsys    = "Machine Learning and Systems (MLSys)"}

@inproceedings{wang2024mmlupro,
  title={{MMLU-Pro: A More Robust and Challenging Multi-Task Language Understanding Benchmark}},
  author={Wang, Yubo and Ma, Xueguang and Zhang, Ge and Ni, Yuansheng and Chandra, Abhranil and others},
  booktitle={The Thirty-eight Conference on Neural Information Processing Systems Datasets and Benchmarks Track},
  year={2024}
}

@inproceedings{rein2024gpqa,
  title={{GPQA: A Graduate-Level Google-Proof Q\&Q Benchmark}},
  author={Rein, David and Hou, Betty Li and Stickland, Asa Cooper and Petty, Jackson and Pang, Richard Yuanzhe and Dirani, Julien and Michael, Julian and Bowman, Samuel R},
  booktitle={First Conference on Language Modeling},
  year={2024}
}

@article{huang2023ceval,
  title={{C-Eval: A Multi-Level Multi-Discipline Chinese Evaluation Suite for Foundation Models}},
  author={Huang, Yuzhen and Bai, Yuzhuo and Zhu, Zhihao and Zhang, Junlei and Zhang, Jinghan and Su, Tangjun and Liu, Junteng and Lv, Chuancheng and Zhang, Yikai and Fu, Yao and others},
  journal={Advances in Neural Information Processing Systems},
  volume={36},
  pages={62991--63010},
  year={2023}
}

@misc{aime2025aime,
  title={{AIME Problems and Solutions}},
  author={{AIME}},
  year={2025},
  url={https://artofproblemsolving.com/wiki/index.php/AIME_Problems_and_Solutions}
}

@article{he2024olympiadbench,
  title={{OlympiadBench: A Challenging Benchmark for Promoting AGI with Olympiad-Level Bilingual Multimodal Scientific Problems}},
  author={He, Chaoqun and Luo, Renjie and Bai, Yuzhuo and Hu, Shengding and Thai, Zhen Leng and Shen, Junhao and Hu, Jinyi and Han, Xu and Huang, Yujie and Zhang, Yuxiang and others},
  journal={arXiv preprint arXiv:2402.14008},
  year={2024}
}

@article{cassano2023multiple,
  title={{MultiPL-E: A Scalable and Polyglot Approach to Benchmarking Neural Code Generation}},
  author={Cassano, Federico and Gouwar, John and Nguyen, Daniel and Nguyen, Sydney and Phipps-Costin, Luna and Pinckney, Donald and Yee, Ming-Ho and Zi, Yangtian and Anderson, Carolyn Jane and Feldman, Molly Q and others},
  journal={IEEE Transactions on Software Engineering},
  volume={49},
  number={7},
  pages={3675--3691},
  year={2023},
  publisher={IEEE}
}

@article{gu2024cruxeval,
  title={{CruxEval: A Benchmark for Code Reasoning, Understanding and Execution}},
  author={Gu, Alex and Rozi{\`e}re, Baptiste and Leather, Hugh and Solar-Lezama, Armando and Synnaeve, Gabriel and Wang, Sida I},
  journal={arXiv preprint arXiv:2401.03065},
  year={2024}
}

@article{zhou2023ifeval,
  title={{Instruction-Following Evaluation for Large Language Models}},
  author={Zhou, Jeffrey and Lu, Tianjian and Mishra, Swaroop and Brahma, Siddhartha and others},
  journal={arXiv preprint arXiv:2311.07911},
  year={2023}
}

@article{dua2019drop,
  title={{DROP: A Reading Comprehension Benchmark Requiring Discrete Reasoning over Paragraphs}},
  author={Dua, Dheeru and Wang, Yizhong and Dasigi, Pradeep and Stanovsky, Gabriel and Singh, Sameer and Gardner, Matt},
  journal={arXiv preprint arXiv:1903.00161},
  year={2019}
}

@article{rajpurkar2018know,
  title={Know what you don't know: Unanswerable questions for SQuAD},
  author={Rajpurkar, Pranav and Jia, Robin and Liang, Percy},
  journal={arXiv preprint arXiv:1806.03822},
  year={2018}
}

@article{ma2024kor,
  title={Kor-bench: Benchmarking language models on knowledge-orthogonal reasoning tasks},
  author={Ma, Kaijing and Du, Xinrun and Wang, Yunran and Zhang, Haoran and Wen, Zhoufutu and Qu, Xingwei and Yang, Jian and Liu, Jiaheng and Liu, Minghao and Yue, Xiang and others},
  journal={arXiv preprint arXiv:2410.06526},
  year={2024}
}

@article{zellers2019hellaswag,
  title={Hellaswag: Can a machine really finish your sentence?},
  author={Zellers, Rowan and Holtzman, Ari and Bisk, Yonatan and Farhadi, Ali and Choi, Yejin},
  journal={arXiv preprint arXiv:1905.07830},
  year={2019}
}

@article{zhuo2024bigcodebench,
  title={Bigcodebench: Benchmarking code generation with diverse function calls and complex instructions},
  author={Zhuo, Terry Yue and Vu, Minh Chien and Chim, Jenny and Hu, Han and Yu, Wenhao and Widyasari, Ratnadira and Yusuf, Imam Nur Bani and Zhan, Haolan and He, Junda and Paul, Indraneil and others},
  journal={arXiv preprint arXiv:2406.15877},
  year={2024}
}

@inproceedings{patil2025bfcl,
title={The Berkeley Function Calling Leaderboard (BFCL): From Tool Use to Agentic Evaluation of Large Language Models}, 
author={Patil, Shishir G. and Mao, Huanzhi and Cheng-Jie Ji, Charlie and Yan, Fanjia and Suresh, Vishnu and Stoica, Ion and E. Gonzalez, Joseph},
booktitle={Forty-second International Conference on Machine Learning},
year={2025},
}

@article{yu2018spider,
  title={Spider: A large-scale human-labeled dataset for complex and cross-domain semantic parsing and text-to-sql task},
  author={Yu, Tao and Zhang, Rui and Yang, Kai and Yasunaga, Michihiro and Wang, Dongxu and Li, Zifan and Ma, James and Li, Irene and Yao, Qingning and Roman, Shanelle and others},
  journal={arXiv preprint arXiv:1809.08887},
  year={2018}
}

@article{qiu2025phybench,
  title={Phybench: Holistic evaluation of physical perception and reasoning in large language models},
  author={Qiu, Shi and Guo, Shaoyang and Song, Zhuo-Yang and Sun, Yunbo and Cai, Zeyu and Wei, Jiashen and Luo, Tianyu and Yin, Yixuan and Zhang, Haoxu and Hu, Yi and others},
  journal={arXiv preprint arXiv:2504.16074},
  year={2025}
}

@article{joshi2017triviaqa,
  title={Triviaqa: A large scale distantly supervised challenge dataset for reading comprehension},
  author={Joshi, Mandar and Choi, Eunsol and Weld, Daniel S and Zettlemoyer, Luke},
  journal={arXiv preprint arXiv:1705.03551},
  year={2017}
}

@inproceedings{suzgun2023challenging,
  title={Challenging big-bench tasks and whether chain-of-thought can solve them},
  author={Suzgun, Mirac and Scales, Nathan and Sch{\"a}rli, Nathanael and Gehrmann, Sebastian and Tay, Yi and Chung, Hyung Won and others},
  booktitle={Findings of the Association for Computational Linguistics: ACL 2023},
  pages={13003--13051},
  year={2023}
}

@inproceedings{kazemi2025big,
  title={Big-bench extra hard},
  author={Kazemi, Mehran and Fatemi, Bahare and Bansal, Hritik and Palowitch, John and Anastasiou, Chrysovalantis and Mehta, Sanket Vaibhav and Jain, Lalit K and Aglietti, Virginia and Jindal, Disha and Chen, Yuanzhu Peter and others},
  booktitle={Proceedings of the 63rd Annual Meeting of the Association for Computational Linguistics (Volume 1: Long Papers)},
  pages={26473--26501},
  year={2025}
}

@article{sprague2023musr,
  title={Musr: Testing the limits of chain-of-thought with multistep soft reasoning},
  author={Sprague, Zayne and Ye, Xi and Bostrom, Kaj and Chaudhuri, Swarat and Durrett, Greg},
  journal={arXiv preprint arXiv:2310.16049},
  year={2023}
}

@article{lin2025zebralogic,
  title={Zebralogic: On the scaling limits of llms for logical reasoning},
  author={Lin, Bill Yuchen and Bras, Ronan Le and Richardson, Kyle and Sabharwal, Ashish and Poovendran, Radha and Clark, Peter and Choi, Yejin},
  journal={arXiv preprint arXiv:2502.01100},
  year={2025}
}

@article{saparov2022language,
  title={Language models are greedy reasoners: A systematic formal analysis of chain-of-thought},
  author={Saparov, Abulhair and He, He},
  journal={arXiv preprint arXiv:2210.01240},
  year={2022}
}

@inproceedings{bisk2020piqa,
  title={Piqa: Reasoning about physical commonsense in natural language},
  author={Bisk, Yonatan and Zellers, Rowan and Gao, Jianfeng and Choi, Yejin and others},
  booktitle={Proceedings of the AAAI conference on artificial intelligence},
  volume={34},
  pages={7432--7439},
  year={2020}
}

@article{hu2020ocnli,
  title={Ocnli: Original chinese natural language inference},
  author={Hu, Hai and Richardson, Kyle and Xu, Liang and Li, Lu and K{\"u}bler, Sandra and Moss, Lawrence S},
  journal={arXiv preprint arXiv:2010.05444},
  year={2020}
}

@article{li2023can,
  title={Can llm already serve as a database interface? a big bench for large-scale database grounded text-to-sqls},
  author={Li, Jinyang and Hui, Binyuan and Qu, Ge and Yang, Jiaxi and Li, Binhua and Li, Bowen and Wang, Bailin and Qin, Bowen and Geng, Ruiying and Huo, Nan and others},
  journal={Advances in Neural Information Processing Systems},
  volume={36},
  pages={42330--42357},
  year={2023}
}

@article{liu2023your,
  title={Is your code generated by chatgpt really correct? rigorous evaluation of large language models for code generation},
  author={Liu, Jiawei and Xia, Chunqiu Steven and Wang, Yuyao and Zhang, Lingming},
  journal={Advances in Neural Information Processing Systems},
  volume={36},
  pages={21558--21572},
  year={2023}
}

@article{gao2024omni,
  title={Omni-math: A universal olympiad level mathematic benchmark for large language models},
  author={Gao, Bofei and Song, Feifan and Yang, Zhe and Cai, Zefan and Miao, Yibo and Dong, Qingxiu and Li, Lei and Ma, Chenghao and Chen, Liang and Xu, Runxin and others},
  journal={arXiv preprint arXiv:2410.07985},
  year={2024}
}

@article{li2024gsm,
  title={Gsm-plus: A comprehensive benchmark for evaluating the robustness of llms as mathematical problem solvers},
  author={Li, Qintong and Cui, Leyang and Zhao, Xueliang and Kong, Lingpeng and Bi, Wei},
  journal={arXiv preprint arXiv:2402.19255},
  year={2024}
}

@article{wei2023cmath,
  title={Cmath: Can your language model pass chinese elementary school math test?},
  author={Wei, Tianwen and Luan, Jian and Liu, Wei and Dong, Shuang and Wang, Bin},
  journal={arXiv preprint arXiv:2306.16636},
  year={2023}
}

@misc{nexusraven,
      title={NexusRaven-V2: Surpassing GPT-4 for Zero-shot Function Calling},
      author={{Nexusflow.ai Team}},
      year={2023},
      url={https://nexusflow.ai/blogs/ravenv2}
}

@misc{opencompass,
      title={open-compass/opencompass},
      author={{Opencompass Team}},
      year={2023},
      url={https://github.com/open-compass/opencompass}
}

@article{jain2024livecodebench,
  title={Livecodebench: Holistic and contamination free evaluation of large language models for code},
  author={Jain, Naman and Han, King and Gu, Alex and Li, Wen-Ding and Yan, Fanjia and Zhang, Tianjun and Wang, Sida and Solar-Lezama, Armando and Sen, Koushik and Stoica, Ion},
  journal={arXiv preprint arXiv:2403.07974},
  year={2024}
}

@misc{song2025seed,
      title={Seed Diffusion: A Large-Scale Diffusion Language Model with High-Speed Inference}, 
      author={Yuxuan Song and Zheng Zhang and Cheng Luo and Pengyang Gao and Fan Xia and Hao Luo and Zheng Li and Yuehang Yang and others},
      year={2025},
      eprint={2508.02193},
      archivePrefix={arXiv},
      primaryClass={cs.CL}, 
}

@article{Blockdiffusion2025,
  title={Block diffusion: Interpolating between autoregressive and diffusion language models},
  author={Arriola, Marianne and Gokaslan, Aaron and Chiu, Justin T and Yang, Zhihan and Qi, Zhixuan and Han, Jiaqi and Sahoo, Subham Sekhar and Kuleshov, Volodymyr},
  journal={arXiv preprint arXiv:2503.09573},
  year={2025}
}

@article{wang2025revolutionizing,
  title={Revolutionizing reinforcement learning framework for diffusion large language models},
  author={Wang, Yinjie and Yang, Ling and Li, Bowen and Tian, Ye and Shen, Ke and Wang, Mengdi},
  journal={arXiv preprint arXiv:2509.06949},
  year={2025}
}

@article{wang2025spg,
  title={SPG: Sandwiched Policy Gradient for Masked Diffusion Language Models},
  author={Wang, Chenyu and Rashidinejad, Paria and Su, DiJia and Jiang, Song and Wang, Sid and Zhao, Siyan and Zhou, Cai and Shen, Shannon Zejiang and Chen, Feiyu and Jaakkola, Tommi and Tian, Yuandong and Liu, Bo},
  journal={arXiv preprint arXiv:2510.09541},
  year={2025}
}

@misc{dfactory,
  title={{dFactory: Easy and Efficient dLLM Fine-Tuning}},
  author={{InclusionAI}},
  year={2025},
  url={https://github.com/inclusionAI/dFactory}
}

@misc{ou_principled_2025,
    title = {Principled {RL} for {Diffusion} {LLMs} {Emerges} from a {Sequence}-{Level} {Perspective}},
    url = {http://arxiv.org/abs/2512.03759},
    doi = {10.48550/arXiv.2512.03759},
    urldate = {2025-12-04},
    publisher = {arXiv},
    author = {Ou, Jingyang and Han, Jiaqi and Xu, Minkai and Xu, Shaoxuan and Xie, Jianwen and Ermon, Stefano and Wu, Yi and Li, Chongxuan},
    month = dec,
    year = {2025},
    note = {arXiv:2512.03759 [cs]},
}

@misc{fu2025areal,
      title={AReaL: A Large-Scale Asynchronous Reinforcement Learning System for Language Reasoning},
      author={Wei Fu and Jiaxuan Gao and Xujie Shen and Chen Zhu and Zhiyu Mei and Chuyi He and Shusheng Xu and Guo Wei and Jun Mei and Jiashu Wang and Tongkai Yang and Binhang Yuan and Yi Wu},
      year={2025},
      eprint={2505.24298},
      archivePrefix={arXiv},
      primaryClass={cs.LG},
      url={https://arxiv.org/abs/2505.24298},
}

@inproceedings{mei2025real,
  author       = {Mei, Zhiyu and Fu, Wei and Li, Kaiwei and Wang, Guangju and Zhang, Huanchen and Wu, Yi},
  title        = {ReaL: Efficient RLHF Training of Large Language Models with Parameter Reallocation},
  booktitle    = {Proceedings of the Eighth Conference on Machine Learning and Systems,
                  MLSys 2025, Santa Clara, CA, USA, May 12-15, 2025},
  publisher    = {mlsys.org},
  year         = {2025},
}

@misc{kang_parallelbench_2025,
    title = {{ParallelBench}: {Understanding} the {Trade}-offs of {Parallel} {Decoding} in {Diffusion} {LLMs}},
    shorttitle = {{ParallelBench}},
    url = {http://arxiv.org/abs/2510.04767},
    doi = {10.48550/arXiv.2510.04767},
    urldate = {2026-02-05},
    publisher = {arXiv},
    author = {Kang, Wonjun and Galim, Kevin and Oh, Seunghyuk and Lee, Minjae and Zeng, Yuchen and Zhang, Shuibai and Hooper, Coleman and Hu, Yuezhou and Koo, Hyung Il and Cho, Nam Ik and Lee, Kangwook},
    month = oct,
    year = {2025},
    note = {arXiv:2510.04767 [cs]},
}

@inproceedings{wang_remasking_2025,
    title = {Remasking {Discrete} {Diffusion} {Models} with {Inference}-{Time} {Scaling}},
    url = {https://openreview.net/forum?id=IJryQAOy0p},
    language = {en},
    urldate = {2026-02-05},
    author = {Wang, Guanghan and Schiff, Yair and Sahoo, Subham Sekhar and Kuleshov, Volodymyr},
    month = oct,
    year = {2025},
}

@misc{lee_effective_2025,
    title = {Effective {Test}-{Time} {Scaling} of {Discrete} {Diffusion} through {Iterative} {Refinement}},
    url = {http://arxiv.org/abs/2511.05562},
    doi = {10.48550/arXiv.2511.05562},
    urldate = {2026-02-05},
    publisher = {arXiv},
    author = {Lee, Sanghyun and Kim, Sunwoo and Kim, Seungryong and Park, Jongho and Park, Dongmin},
    month = nov,
    year = {2025},
    note = {arXiv:2511.05562 [cs]},
}

@inproceedings{rutte_generalized_2025,
    title = {Generalized {Interpolating} {Discrete} {Diffusion}},
    url = {https://openreview.net/forum?id=rvZv7sDPV9},
    language = {en},
    urldate = {2026-02-05},
    author = {Rütte, Dimitri von and Fluri, Janis and Ding, Yuhui and Orvieto, Antonio and Schölkopf, Bernhard and others},
    month = jun,
    year = {2025},
}

@misc{antgroupdeepxputeamPowerDiffusionLLMs,
  title = {Power {{Up Diffusion LLMs}}: {{Day}}-0 {{Support}} for {{LLaDA}} 2.0 \textbar{} {{LMSYS Org}}},
  shorttitle = {Power {{Up Diffusion LLMs}}},
  author = {{Ant Group Team} and {SGLang Team}},
  urldate = {2026-02-05},
  langid = {english},
  url= {https://lmsys.org/blog/2025-12-19-diffusion-llm}
}

@misc{alphamoe,
  title = {{Alpha-MoE}: A Megakernel for Faster Tensor Parallel Inference
},
  author = {Aleph-Alpha},
  urldate = {2026-02-05},
  url= {https://aleph-alpha.com/alpha-moe-a-megakernel-for-faster-tensor-parallel-inference/}
}

@article{ma2025dinfer,
  title={dinfer: An efficient inference framework for diffusion language models},
  author={Ma, Yuxin and Du, Lun and Wei, Lanning and Chen, Kun and Xu, Qian and Wang, Kangyu and Feng, Guofeng and Lu, Guoshan and Liu, Lin and Qi, Xiaojing and others},
  journal={arXiv preprint arXiv:2510.08666},
  year={2025}
}

@misc{lingteam2025stepevolvesscalingreinforcement,
      title={Every Step Evolves: Scaling Reinforcement Learning for Trillion-Scale Thinking Model}, 
      author={{Ling Team} and others},
      year={2025},
      eprint={2510.18855},
      archivePrefix={arXiv},
      primaryClass={cs.CL},
      url={https://arxiv.org/abs/2510.18855}, 
}

\clearpage

\end{document}